\documentclass{article} 
\usepackage[final]{colm2026_conference}
\usepackage[OT1]{fontenc}

\usepackage{microtype}
\usepackage{url}
\usepackage{booktabs}
\usepackage{array}
\usepackage{graphicx}
\usepackage{amsmath,amssymb,bm}
\usepackage{algorithm}
\usepackage{algpseudocode}
\usepackage{hyperref}

\definecolor{darkblue}{rgb}{0, 0, 0.5}
\hypersetup{colorlinks=true, citecolor=darkblue, linkcolor=darkblue, urlcolor=darkblue,
  pdftitle={FOCUS: Closed-Loop Attention Feedback for Efficient Vision-Language Understanding},
  pdfauthor={Akul Santhosh and Aastha Jhunjhunwala}}

\algrenewcommand\algorithmicrequire{\textbf{Input:}}

\title{FOCUS: Closed-Loop Attention Feedback for Efficient Vision-Language Understanding}

\author{Akul Santhosh \\
NVIDIA \\
\texttt{akuls@nvidia.com} \\
\And
Aastha Jhunjhunwala \\
NVIDIA \\
\texttt{aasthaj@nvidia.com}
}

\begin{document}

\maketitle
\AddToShipoutPictureFG*{\AtPageLowerLeft{\put(108,12){%
  \footnotesize Code: \url{https://github.com/akul-nv/focus}}}}

\begin{abstract}
Processing every video frame through Vision-Language Models (VLMs) adds
significant inference cost and limits their ability to comprehend real-time
video. Existing approaches either sub-sample frames uniformly, ignoring content
and missing important events, or prune visual tokens inside the VLM to cut
per-call cost while still invoking the model on every sampled frame. We introduce
FOCUS, a training-free three-layer framework that reduces the number of
invocations and steers each one towards what matters. Temporal gating measures
inter-frame dissimilarity in the
patch-embedding space of a smaller VLM from the same family, triggering burst captures only when the scene changes in a
semantically meaningful way. Patch-level dissimilarity is then accumulated across the burst
and fused with a top-down attention prior into a spatial focus map that encodes
where motion and semantic relevance coincide. That map calibrates the VLM's value
cache during generation, progressively amplifying vision tokens in task-relevant
regions and attenuating off-target ones, without removing any token from the
context. The loop closes after each response: text-to-vision cross-attention is
extracted from the decoder and fed back as the prior for the next burst, letting
the system converge on semantically important regions over successive
observations. We evaluate FOCUS on the
500-video EgoSchema subset in streaming mode, with a single Qwen3-VL-8B
backbone shared by every method. It reaches 74.8\%
accuracy using 25.4 VLM calls per clip, against 74.6\% with 55 calls for uniform
sampling and 74.2\% with 55 calls for BOLT. On egocentric video captioning across
four domains, it matches the best baseline's quality while issuing 57\% fewer VLM
invocations than uniform sampling.
\end{abstract}


\section{Introduction}
\label{sec:introduction}

Vision-Language Models (VLMs) can now describe scenes, answer visual questions, and reason spatially across images and videos~\citep{liu2023llava, wang2024qwen2vl, maaz2024videochatgpt, zhang2024llavevideo, li2025llavaonevision}.
Applying them to video remains expensive: a 30-fps video lasting one minute produces 1,800 frames, each adding hundreds of visual tokens to the model's context.  Where video arrives as a continuous stream, as from an egocentric camera on a robot or a wearable device narrating a user's surroundings, the full clip is never available in advance and the system must decide what to process as frames arrive.  Existing VLM pipelines are not designed for this setting: they either assume access to the complete video or allocate compute uniformly across frames regardless of content.

The most common mitigation is uniform temporal sub-sampling: selecting every $n$-th frame under the assumption that intermediate frames carry little new information~\citep{maaz2024videochatgpt, zhang2024llavevideo}.  It is blind to content, allocating the same compute budget to static, repetitive segments as to brief, information-dense events, and risking the loss of critical transitions.  A second family reduces the cost of each VLM call rather than the number of calls: token pruning methods~\citep{chen2024fastv, shang2025prumerge} drop vision tokens that contribute little to the output, while temporal compression methods~\citep{tao2025dycoke, bagrov2025evs} merge redundant tokens across consecutive frames, both lowering per-call cost but still invoking the model on every sampled frame.  A third line uses adaptive frame selection, where an agent or retrieval model identifies which frames to process~\citep{wang2024videoagent, yu2025framevoyager, buch2025flexframe}; these can substantially reduce VLM calls, but typically require multiple selection rounds, an auxiliary retrieval model, or access to the full video, making them unsuitable for streaming.

Our work differs from all three categories in two ways.  First, FOCUS reduces the \emph{total number} of VLM calls through content-adaptive gating: it measures inter-frame dissimilarity in a VLM patch-embedding space rather than in pixel space, and triggers inference only when the scene changes meaningfully.  Second, it closes the loop between perception and understanding: after each VLM call the decoder's attention is fed back as a spatial prior for the next observation, so the system progressively converges on task-relevant regions without gradient updates or learned parameters (Section~\ref{sec:method}).  A secondary effect is that the VLM generates more concise outputs once scene context is established, cutting per-call inference time by up to 41\% on individual bursts (Section~\ref{sec:qualitative}).

We evaluate FOCUS on two complementary benchmarks. The primary one is the 500-video
EgoSchema~\citep{mangalam2023egoschema} subset, where every method runs on one Qwen3-VL-8B
backbone at a matched cadence and answers are graded against ground truth. FOCUS
reaches 74.8\% accuracy with 25.4 VLM calls per three-minute clip, against 74.6\% with 55
calls for uniform sampling, 74.2\% with 55 calls for BOLT~\citep{liu2025bolt} and 71.4\% with
12 calls for DyCoke~\citep{tao2025dycoke}. It therefore matches the strongest baselines at under
half their call budget.

The second benchmark is a fine-grained egocentric captioning study across four domains,
25 videos scored per criterion by GPT-5.2. Here FOCUS attains the best overall quality
(3.92/5) while issuing 57\% fewer VLM invocations than uniform sampling and 16\% fewer
than DyCoke and BOLT. The gaps between methods are small and the sample is limited,
so we read this benchmark as a per-criterion diagnostic rather than as evidence of a quality
improvement. The claim we make throughout is matched-or-better quality at substantially
lower VLM cost, in a streaming setting applicable to robotic manipulation, accessibility
narration and live event summarization.

\paragraph{Contributions.}
\begin{enumerate}
\itemsep0pt
\item Content-adaptive temporal gating in a VLM patch-embedding space, substantially reducing VLM invocations while maintaining quality.
  \item A spatial focus pipeline that fuses bottom-up patch dissimilarity with top-down attention priors to steer each VLM call toward task-relevant regions.
  \item Value-cache calibration, which scales vision-token values in the decoder's KV cache using that prior, amplifying relevant regions without discarding context.
  \item A closed-loop feedback architecture that feeds decoder attention back into the spatial prior, enabling progressive refinement across successive observations.
  \item A controlled streaming evaluation in which all methods share one backbone and cadence, plus component, threshold and per-domain ablations separating the compute contribution of gating from the quality contribution of calibration.
\end{enumerate}


\section{Related Work}
\label{sec:related}

\paragraph{Frame selection for video VLMs.}
Uniform temporal sampling remains the default strategy for feeding video to VLMs, but it ignores content entirely and risks missing transient events.
Content-aware alternatives have emerged along two axes.
Query-guided methods such as VideoAgent~\citep{wang2024videoagent} use an LLM to iteratively retrieve frames by caption similarity, achieving strong accuracy at the cost of multiple VLM and LLM calls per video. Training-based selectors such as Frame-Voyager~\citep{yu2025framevoyager} and the Flexible Frame Selector~\citep{buch2025flexframe} learn to pick frame subsets instead, but require task-specific supervision.
BOLT~\citep{liu2025bolt} avoids training altogether by scoring frames via CLIP similarity to the query, though it must encode and score all frames upfront and assumes a predefined query is available.
All of these methods focus on improving which frames the VLM sees.

\paragraph{Visual token reduction.}
Rather than reducing the number of VLM invocations, a large body of work reduces the number of vision tokens \textit{within} each invocation.
FastV~\citep{chen2024fastv}, SparseVLM~\citep{zhang2025sparsevlm}, LLaVA-PruMerge~\citep{shang2025prumerge} and PyramidDrop~\citep{xing2025pyramiddrop} prune or merge vision tokens at various stages of the model, and video-specific extensions such as PruneVid~\citep{huang2025prunevid} and HoliTom~\citep{shao2025holitom} additionally exploit temporal redundancy across frames. These approaches typically remove a fraction of the tokens, yielding modest speedups while preserving most of the original accuracy.

\paragraph{KV cache optimization and generation steering.}
A related family operates directly on the decoder's KV cache.
DyCoke~\citep{tao2025dycoke} combines temporal token merging at prefill with dynamic KV cache pruning during decoding, evicting less-attended vision-token entries, and VL-Cache~\citep{tu2025vlcache} introduces modality-aware cache budget allocation exploiting VLM-specific sparsity; both reduce memory and compute by permanently discarding cache entries.
SPARC~\citep{jung2025sparc} instead shows that selectively \emph{scaling} value representations during decoding steers VLM generation without modifying the input, using EMA-smoothed attention ratios computed dynamically.
That is the mechanism we build on in Section~\ref{sec:value_cache}; our departure is that the scaling factors come from an explicit spatial prior carried across bursts rather than from token rankings computed within a single generation.

\noindent
To our knowledge, FOCUS is the first training-free framework to unify content-adaptive frame selection, spatial attention steering and generation-time value-cache calibration in one closed-loop system that applies naturally to streaming video.  Our baselines span these axes under a single backbone: uniform sampling as the non-adaptive upper bound on calls, BOLT as a training-free query-aware selector, and DyCoke as a KV-cache and token-merging method (Section~\ref{sec:setup}).


\section{Method}
\label{sec:method}

FOCUS is a three-layer system for efficient, spatially focused VLM inference on streaming video. It asks: \textit{(1)} should this frame be processed at all? \textit{(2)} where in the frame should the VLM focus? \textit{(3)} how can the VLM's internal understanding improve future focus? These correspond to \textit{semantic frame gating} (Section~\ref{sec:frame_gating}), \textit{value-cache calibration with spatial priors} (Section~\ref{sec:value_cache}) and \textit{closed-loop attention feedback} (Section~\ref{sec:attention_feedback}). Temporal gating and spatial attention have been explored independently; the central contribution of FOCUS is closing the loop between them, extracting the VLM's own decoder attention after each invocation and feeding it back as the spatial prior for the next, steering calibration across bursts without gradient updates or learned parameters.

Appendix~\ref{app:notation} collects the notation in a single table and Appendix~\ref{app:algorithm} gives the full procedure as pseudocode.

\begin{figure}[t]
    \centering
    \includegraphics[width=0.40\linewidth]{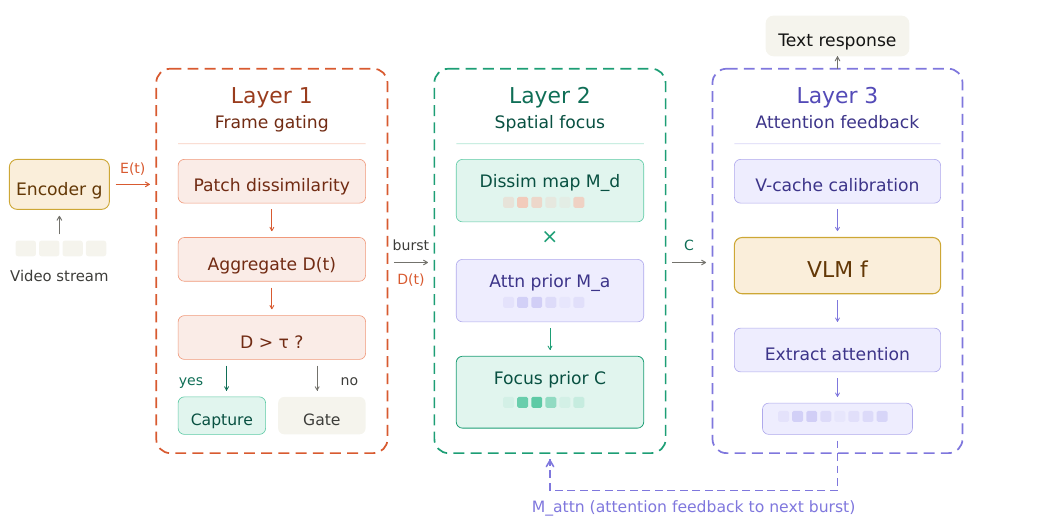}
    \caption{Architecture of the three-layer FOCUS system.}
    \label{fig:pipeline}
\end{figure}

\subsection{Semantic Frame Gating}
\label{sec:frame_gating}

\paragraph{Dissimilarity in embedding space.}
Any streaming system must decide which frames deserve expensive processing. Traditional approaches make that decision from pixel-level cues such as histogram differences, optical flow or hand-crafted features~\citep{horn1981opticalflow, swain1991color}. The problem is that these signals are decoupled from the model that will ultimately interpret the frame. FOCUS instead measures change in the kind of representation the VLM itself works in. A gating encoder $g$ maps each arriving frame $t$ to patch embeddings $\mathbf{e}_i^{(t)}$ over a grid of $P$ patches, and we compare them against the previously processed frame $\mathrm{ref}$, which is updated after every step, by cosine dissimilarity $d_i^{(t)} = 1 - \cos(\mathbf{e}_i^{(t)}, \mathbf{e}_i^{(\mathrm{ref})})$. Averaging over patches gives one scalar per frame, $D^{(t)} = \tfrac{1}{P}\sum_i d_i^{(t)}$, which trades off sensitivity to localized against global change. Because these are learned semantic features rather than pixels, a meaningful change such as a hand entering the scene produces a large $D^{(t)}$ even when it occupies few pixels, while a visually dominant but semantically vacuous one such as camera shake produces a small one.\footnote{The gating encoder $g$ is a smaller model from the same VLM family (Qwen2-VL-2B) than the one that generates captions and answers (Qwen3-VL-8B), because gating runs on every arriving frame and must keep pace with the stream. The two share the same ViT design and patch tokenization, so gating is computed in a \emph{compatible} patch-embedding space within one model family; we do \emph{not} claim that gating and generation share one set of weights. Gating requires only that $g$ preserve semantic patch similarity.}

\paragraph{Gating state machine.}
The purpose of gating is to reduce VLM computation by skipping frames that carry no new information. The logic is a two-state machine over \textsc{capture} and \textsc{watch}. In \textsc{capture} the system collects a burst of $B$ consecutive frames, which gives the VLM a short temporal window to reason about an action rather than a single snapshot. Once the burst is complete the system moves to \textsc{watch} and frames are skipped. It returns to \textsc{capture} when dissimilarity exceeds the threshold, $D^{(t)} > \tau$, or when a safety timeout of $N_{\max}$ consecutively skipped frames is reached, which ensures the VLM is never starved of input during a prolonged static scene. The frame that triggers the transition becomes the first frame of the new burst.

\paragraph{Efficiency metric.}
To measure the compute savings from gating we report $\rho$, the ratio of total VLM inference time to video duration:
\begin{equation}
\label{eq:efficiency}
    \rho = \frac{\sum_{n} t_{\mathrm{vlm}}^{(n)}}{T_{\mathrm{video}}} .
\end{equation}
A ratio $\rho < 1$ means the VLM keeps pace with the stream faster than real time. The metric covers VLM inference alone and excludes the gating encoder, which costs $t_{\mathrm{enc}} = 34$\,ms per frame regardless of what the frame contains or which gating policy is in force; leaving out a constant isolates the one quantity a method can actually change. That cost is still a real charge against a deployment, and FOCUS pays it on every arriving frame where uniform sampling and DyCoke do not. Adding it back is a single multiplication, $t_{\mathrm{enc}} N_{\mathrm{frames}} / T_{\mathrm{video}}$, worth $0.77$ on our captioning benchmark against $\rho = 0.59$ for FOCUS. In a deployed system the two stages overlap, since neither waits on the other within a frame, so the cost is the larger of the two rather than their sum. Once encoding keeps pace with the stream it becomes a floor, and VLM inference is the only term left for any method to improve.

\subsection{Value-Cache Calibration with Spatial Priors}
\label{sec:value_cache}

\begin{figure*}[t]
    \centering
    \includegraphics[width=0.60\textwidth]{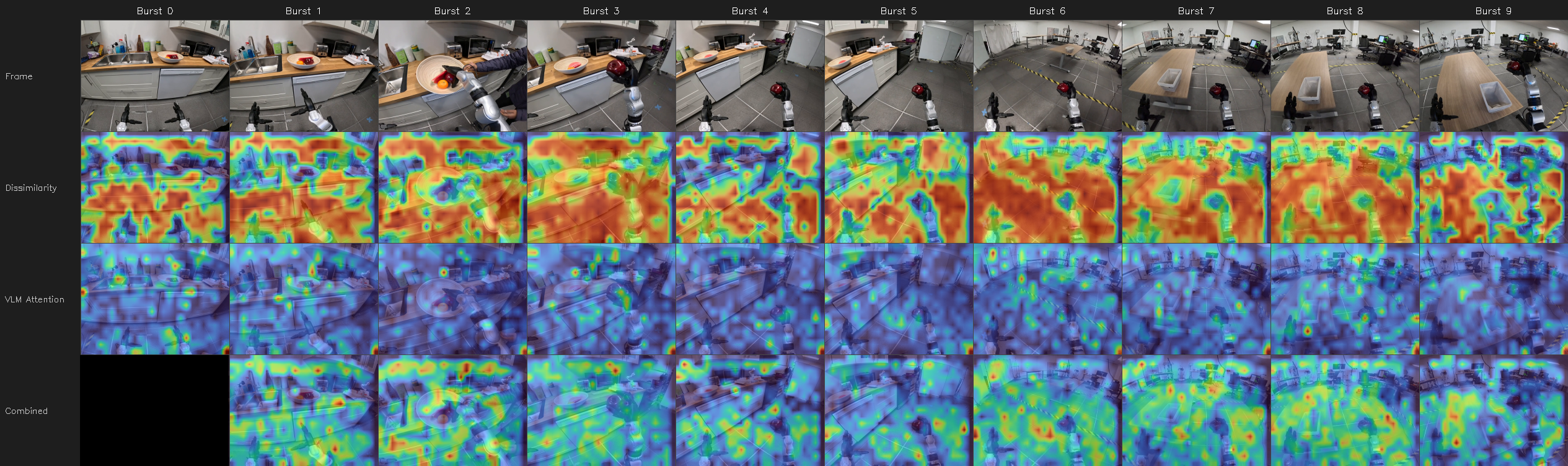}
    \caption{Spatial prior evolution across bursts (robotic teleoperation). Columns are bursts; rows are a representative frame, the dissimilarity map, the decoder attention after generation, and the combined prior (absent at Burst 0, the orienting phase).}
    \label{fig:convergence}
\end{figure*}

Frame gating decides when to invoke the VLM. We observe that additionally modulating where the VLM attends within each invocation improves task performance across all settings (Section~\ref{sec:experiments}). The spatial prior that does this steering tends to concentrate on regions of activity, such as a robot arm or the object it manipulates (Figure~\ref{fig:convergence}), even though nothing in the mechanism supervises which region to pick.

\paragraph{Spatial prior construction.}
The per-patch dissimilarity vector $\mathbf{d}^{(t)}$ is already a byproduct of gating, and it carries spatial information the gating decision throws away: it encodes where in the frame the change occurred. We reshape it onto the encoder's grid $\mathbf{S}^{(t)} \in \mathbb{R}^{H_g \times W_g}$ and take the element-wise maximum over the $B$ frames of the burst, $\mathbf{M}_{\mathrm{dissim}} = \mathrm{norm}(\max_{t} \mathbf{S}^{(t)})$, where $\mathrm{norm}(\cdot)$ rescales to $[0,1]$. Taking the maximum rather than the mean retains brief but significant motions at full strength even when they occupy only a few frames of the burst.

This map captures physical change, but not all change is task-relevant. A flickering monitor produces high dissimilarity while contributing nothing to the task, while a stationary object the VLM has been attending to, such as a tool about to be grasped, may show almost no dissimilarity yet be exactly what matters. Neither signal suffices alone, so we fuse this bottom-up map with a top-down attention prior $\mathbf{M}_{\mathrm{attn}}$ from the previous burst (Section~\ref{sec:attention_feedback}):

\begin{equation}
\label{eq:focus_prior}
    \mathbf{C} = \mathrm{norm}\!\left(\mathbf{M}_{\mathrm{dissim}} \odot \left(\varepsilon + \mathbf{M}_{\mathrm{attn}}^{\gamma}\right)\right),
\end{equation}
where $\gamma$ (default $0.5$) controls how sharply attention modulates dissimilarity and $\varepsilon = 0.01$ is a floor ensuring low-attention regions are down-weighted but never zeroed. The key property is that a region must score highly on \textit{both} signals to receive a strong prior.

\paragraph{Value-cache scaling.}
SPARC~\citep{jung2025sparc} showed that selectively scaling internal representations steers VLM behaviour without modifying the input. We build on this but depart from it: where SPARC selects tokens dynamically via EMA-smoothed attention ratios, we apply the fixed spatial prior of Eq.~\ref{eq:focus_prior}, flattened and mapped directly onto the vision-token positions $\mathcal{V}$ (resampled by nearest-neighbour interpolation when dynamic-resolution tiling makes the lengths differ), so calibration operates in spatial coordinates rather than token-ranking space. Each token $j \in \mathcal{V}$ receives a prior $C_j \in [0,1]$ and a per-step scale
\begin{equation}
\label{eq:value_scaling}
s_j =
\begin{cases}
1 + (\alpha_f - 1)\, C_j, & C_j \in [r_{\mathrm{lo}}, r_{\mathrm{hi}}] \quad\text{(amplified)},\\[1pt]
\max\!\big(1 - \beta\,(1 - C_j),\ 0.5\big), & C_j \notin [r_{\mathrm{lo}}, r_{\mathrm{hi}}] \quad\text{(suppressed)},
\end{cases}
\end{equation}
applied to the value cache of every calibrated layer at every autoregressive decoding step:
\begin{equation}
\label{eq:value_update}
\mathbf{V}^{(\ell)}_j \leftarrow s_j \cdot \mathbf{V}^{(\ell)}_j
\qquad \forall\, j \in \mathcal{V},\ \ \forall\, \ell \in [l_{\mathrm{start}},\, l_{\mathrm{end}}].
\end{equation}
Here $\alpha_f$ is the focus strength (default $1.05$), $\beta$ the suppression strength (default $0.04$), and $[r_{\mathrm{lo}}, r_{\mathrm{hi}}]$ the focus range (Section~\ref{sec:setup}); $\mathbf{V}^{(\ell)}_j$ is the cached value of position $j$ at decoder layer $\ell$, and unless stated otherwise we calibrate the full stack of $L = 28$ layers. Two properties of this update are worth stating plainly. The floor of $0.5$ bounds the \emph{per-step} factor only, not the compounded product, and at $\beta = 0.04$ it never binds, since the smallest per-step factor is $0.96$. And because the update multiplies a cache that persists across the whole response, the factors compound: after $N$ decode steps the effective multiplier is $s_j^N$, so a per-step nudge of $1.05$ becomes a scaling of ${\approx}131\times$ by our 100-token cap while a fully off-prior token falls to ${\approx}0.017\times$. Appendix~\ref{app:hparams} gives the remaining implementation details.

This works because value vectors determine how much each token contributes to the attention-weighted sum during generation. Scaling them amplifies the information flowing from the prior's regions into every subsequently generated token. Keys and queries are never touched, so the attention weights are exactly those of the uncalibrated model and the next sub-layer's normalisation absorbs any common scale; what survives is the relative re-weighting among vision tokens, and no token leaves the context. Prompt-level alternatives such as region-of-interest cropping, textual spatial hints and soft foveation all degraded performance in preliminary experiments, apparently because they discard context the model needs for coherent reasoning.

\subsection{Closed-Loop Attention Feedback}
\label{sec:attention_feedback}

Most frame selection systems are purely feed-forward: the selector never receives feedback from the model that consumes its output. FOCUS closes this loop by extracting the VLM's internal attention after each burst and feeding it back as the top-down prior $\mathbf{M}_{\mathrm{attn}}$ for the next.

\paragraph{Attention extraction.}
Once the VLM has answered for burst $n$, we want a spatial map of where it actually looked while generating that answer. A single additional forward pass over the prompt and the generated response, with attention outputs enabled, yields the attention tensors $\mathbf{A}^{(\ell,h)} \in \mathbb{R}^{S \times S}$ for every decoder layer $\ell$ and head $h$ over the length-$S$ sequence. We aggregate the last $N_a$ layers (default $4$), since later decoder layers carry more task-specific attention~\citep{clark2019bert_attention}, by max-pooling across heads and averaging across layers. Max-pooling preserves the most spatially selective head rather than diluting it against heads that attend almost entirely to text:

\begin{equation}
\label{eq:attention_aggregation}
    \bar{\mathbf{A}} = \frac{1}{N_a} \sum_{\ell=L-N_a}^{L-1} \max_{h} \mathbf{A}^{(\ell,h)} \in \mathbb{R}^{S \times S}.
\end{equation}
From $\bar{\mathbf{A}}$ we keep only the text-to-vision block: how much the generated text attended to each vision token. Text positions before the vision block are dropped, because the causal mask gives them exactly zero attention to it. Collapsing that block to one score per vision token is not a plain average, since generated tokens differ widely in how much they look at the image at all; we weight each text token by the total attention it directs at the image, so tokens that are largely reasoning over text contribute little. The resulting scores are folded back onto the patch grid, averaged over the frames of the burst and normalised to $[0,1]$ to give $\mathbf{M}_{\mathrm{attn}}$. The extra pass has no autoregressive loop and is small relative to the generation it follows. Appendix~\ref{app:attention} states all six steps precisely.

\paragraph{Feedback loop lifecycle.}
After each burst the dissimilarity maps are cleared but $\mathbf{M}_{\mathrm{attn}}$ is retained, so the VLM's understanding from burst $n$ shapes the prior of burst $n+1$. On the first burst no prior exists and the VLM processes the full scene without calibration, an orienting phase; from burst $2$ onward the stored prior is fused with the new dissimilarity map (Eq.~\ref{eq:focus_prior}) and the resulting $\mathbf{C}$ calibrates the value cache (Eq.~\ref{eq:value_update}), each burst producing a new attention map that replaces the previous one (Figure~\ref{fig:convergence}). Alongside it, a rolling window of the last $W$ responses (default $W{=}5$) carries forward as textual context, discouraging repeated descriptions across bursts; this window is enabled identically for every method we compare and every FOCUS ablation, so it is a controlled constant rather than a source of the differences we report.


\section{Experiments}
\label{sec:experiments}

\subsection{Setup}
\label{sec:setup}

\paragraph{Models and configuration.}
FOCUS uses the Qwen2-VL-2B-Instruct~\citep{wang2024qwen2vl} vision encoder for gating and Qwen3-VL-8B-Instruct~\citep{qwen3vl2025} as the inference VLM; the same Qwen3-VL-8B backbone is used for \emph{every} method in every table below. Experiments run in bfloat16 on NVIDIA A100 80\,GB and RTX PRO 6000 GPUs, with all $\rho$ values measured on one RTX PRO 6000. Unless stated otherwise we use a single configuration for every FOCUS experiment: $\tau = 0.4$, $B = 32$, $\alpha_f = 1.05$, $\beta = 0.04$, $[r_{\mathrm{lo}}, r_{\mathrm{hi}}] = [0.75, 1.0]$, greedy decoding and a 100-token cap. Every default was fixed once on a development set disjoint from all evaluation videos and never re-selected afterwards; Appendix~\ref{app:hparams} gives that protocol and tabulates all twelve constants with their roles and sensitivity.

\paragraph{Benchmarks and baselines.}
Our primary benchmark is \textbf{EgoSchema}~\citep{mangalam2023egoschema}, whose publicly released 500-clip subset pairs three-minute egocentric clips with five-way multiple-choice questions and ground-truth answers, and is the split prior systems report on. We complement it with \textbf{egocentric captioning} over 25 videos from Ego4D~\citep{grauman2022ego4d}, EPIC-Kitchens~\citep{damen2022epic}, Ego-Exo4D~\citep{grauman2024egoexo4d} and GR-1 teleoperation recordings~\citep{wu2024gr1} plus one in-house recording, treated as a per-criterion diagnostic: at $n{=}25$ with an LLM judge, gaps are within noise. Our baselines span the design axes of Section~\ref{sec:related} under one backbone: \textbf{uniform sampling}, the non-adaptive upper bound, firing a 32-frame burst on a fixed schedule; \textbf{DyCoke}~\citep{tao2025dycoke}; \textbf{BOLT}~\citep{liu2025bolt}; and \textbf{FOCUS (gating only)}, our gating layer with per-burst captioning but no spatial calibration or feedback, isolating gating from everything downstream. For captioning, GPT-5.2 rates each caption from 1 to 5 on \emph{Actor}, \emph{Activity}, \emph{Object} and \emph{No Hallucination}, with \emph{Overall} their mean~\citep{zheng2023judging}; see Appendix~\ref{app:judge}. Appendices~\ref{sec:ablations_main}--\ref{sec:ablations_appendix} give the component, threshold, per-domain and burst-size ablations.


\subsection{Main Results: EgoSchema under a Matched Backbone}
\label{sec:egoschema}

Every method runs on the 500-video subset at a matched $3.3$\,s cadence with matched prompting, so only the selection and steering policy varies. The setting is \emph{streaming}: frames arrive in order, with no future frames and no pre-specified query. FOCUS passes a cumulative narrative built from its burst captions, with the five answer choices, to the VLM in one final QA call; the baselines capture a 32-frame burst per interval and aggregate their outputs for that call.

\begin{table}[t]
\centering
\caption{EgoSchema subset accuracy (500 videos, 5-way multiple choice) in the
streaming setting.  All methods share the Qwen3-VL-8B backbone at a matched
$3.3$\,s cadence; accuracy is the mean over five runs, ``VLM Calls'' the mean
invocations per three-minute clip.  The two FOCUS rows issue identical calls by
construction: gating alone determines how many bursts fire.}
\label{tab:egoschema}
\smallskip
\small
\begin{tabular}{@{}lcc@{}}
\toprule
\textbf{Method (streaming, matched cadence)} & \textbf{Accuracy} & \textbf{VLM Calls} \\
\midrule
Uniform (streaming)                       & 74.6\% & 55 \\
BOLT~\citep{liu2025bolt} (streaming)      & 74.2\% & 55 \\
DyCoke~\citep{tao2025dycoke} (streaming)  & 71.4\% & \underline{12} \\
\midrule
FOCUS (gating only)              & 73.6\% & 25.4 \\
FOCUS (w/ Feedback)              & \textbf{74.8\%} & 25.4 \\
\bottomrule
\end{tabular}
\end{table}

FOCUS with feedback reaches $74.8\%$ using $25.4$ VLM calls per clip, against $74.6\%$ for uniform sampling and $74.2\%$ for BOLT, both of which spend $55$. The accuracy differences here are not the result. One video is worth $0.2$ points on a 500-item benchmark, so those two gaps come to one video and three, and five runs cannot resolve differences that small; we claim no accuracy win. What the table separates cleanly is the call budget, where the gap is large and deterministic. DyCoke is the informative exception: it issues the fewest calls but gives up $3.4$ points, the price of compressing each call rather than deciding whether to make it. The $1.2$ points between our own two rows sit inside the same band, so the case for feedback rests on the object-grounding effect isolated in Appendix~\ref{sec:ablations_main} rather than on this number.

That the two FOCUS rows issue identical calls is structural, not a coincidence. Gating (Layer~1) sits upstream of calibration and feedback (Layers~2--3) and is independent of them, so the burst count is fixed by $D^{(t)} > \tau$ and $B$ alone, measured at $25.4$ per clip (sd $6.5$ across clips).\footnote{An earlier version of this comparison reported $19$ and $26$ calls for these rows in error; the shared $25.4$ supersedes them.} Feedback does not change how many calls are made, only what each one costs: it adds one attention-extraction pass per burst, moving $\rho$ from $0.57$ to $0.59$ on captioning (Table~\ref{tab:main_results}). One caveat: at 6~fps most arriving frames fall inside a burst, so gating here buys fewer, longer-gap bursts rather than frame-level skipping, leaving the call-count saving untouched since an invocation is billed per burst (Appendix~\ref{sec:ablations_appendix}).

Prior systems on this subset use much larger proprietary backbones and non-streaming protocols, so we leave them out of Table~\ref{tab:egoschema}: those differences would confound exactly the comparison it isolates. Appendices~\ref{app:offline} and~\ref{app:egoschema_breakdown} add an offline sanity check and a per-answer breakdown.


\subsection{Egocentric Video Captioning}
\label{sec:main_results}

\begin{table}[t]
\centering
\caption{Captioning quality (GPT-5.2, 1--5) and efficiency on 25 egocentric videos.
$\rho$ is the VLM compute ratio (Eq.~\ref{eq:efficiency}); lower is better. Same
backbone throughout, but not decoding-matched: DyCoke and BOLT use sampled decoding and
a 200-frame interval before their own selection, against greedy decoding and 100 for
Uniform and FOCUS.  A further reason to read the small quality gaps cautiously.}
\label{tab:main_results}
\smallskip
\small
\setlength{\tabcolsep}{4.2pt}
\begin{tabular}{@{}lccccccc@{}}
\toprule
\textbf{Method} & \textbf{Actor} & \textbf{Activity} & \textbf{Object} & \textbf{No Halluc.} & \textbf{Overall} & $\boldsymbol{\rho}$ & \textbf{VLM Calls} \\
\midrule
Uniform Sampling   & 4.96 & \textbf{4.12} & 4.00 & 2.16 & 3.81 & 4.14 & 516 \\
DyCoke             & 4.80 & 3.88 & 4.04 & 2.52 & 3.81 & 3.07 & 263 \\
BOLT               & 4.76 & \textbf{4.12} & 4.08 & \textbf{2.68} & 3.91 & 0.87 & 263 \\
\midrule
FOCUS (gating only)  & \textbf{5.00} & 4.00 & \textbf{4.12} & 2.44 & 3.89 & \underline{0.57} & \underline{222} \\
FOCUS (w/ Feedback)  & \textbf{5.00} & 3.96 & \textbf{4.12} & 2.60 & \textbf{3.92} & 0.59 & \underline{222} \\
\bottomrule
\end{tabular}
\end{table}

Table~\ref{tab:main_results} is a statement about cost rather than quality. FOCUS with feedback issues 57\% fewer VLM calls than uniform sampling (222 against 516) and 16\% fewer than BOLT and DyCoke, and it runs at $\rho = 0.59$ against BOLT's $0.87$ while satisfying a streaming constraint BOLT does not: BOLT needs every frame upfront and a predefined query. On quality the methods are tied. The $0.01$ margin over BOLT gives $p = .895$ under a paired Wilcoxon test on the 25 matched per-video scores, and the $0.11$ over DyCoke gives $p = .158$, so we claim no quality gain (Appendix~\ref{app:judge}). BOLT's $\rho$ also excludes the CLIP encoding it runs over every frame upfront, so on that axis the comparison is conservative in its favour.

The per-criterion columns are what this benchmark is for. Uniform sampling ties for the best Activity score ($4.12$, with BOLT) yet ranks last on No Hallucination ($2.16$); seeing every frame does not reduce hallucination and appears to increase it, plausibly by flooding the model with redundant visual context. BOLT posts the best No Hallucination score ($2.68$), likely for the mirror-image reason: its CLIP-based selection discards the ambiguous frames that would prompt the model to invent detail. Feedback contributes $+0.03$ overall over gating alone, driven by $+0.16$ on No Hallucination, consistent with the prior suppressing irrelevant content in later bursts, though at $n{=}25$ this is directional rather than established.

\subsection{Qualitative Analysis}
\label{sec:qualitative}
\label{app:qualitative}

What the closed loop does is easiest to see burst by burst. Figure~\ref{fig:convergence} tracks the prior across a teleoperation clip: absent at the orienting burst, then contracting onto the arm and the object it manipulates. Two EPIC-Kitchens clips show the effect on the text.

\begin{figure}[h]
    \centering
    \includegraphics[width=0.8\linewidth]{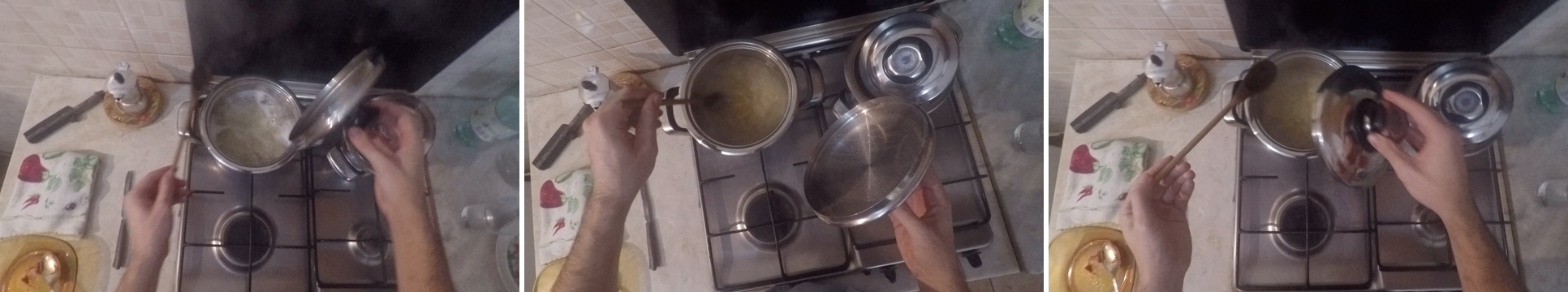}
    \caption{Frames from the soup-stirring clip (EPIC-Kitchens, 15s).}
    \label{fig:soup_frames}
\end{figure}

\paragraph{Attention feedback corrects spatial hallucinations.}
In a 15-second clip of a person stirring soup and covering a pot (Figure~\ref{fig:soup_frames}), both variants select the same two bursts and give identical first-burst descriptions, since the orienting phase has no prior, but their second-burst descriptions diverge:

\begin{quote}
\small
\textbf{Without feedback:}
``\textit{placed the metal lid onto the \underline{right-hand burner} of the stove, covering the \underline{empty pot}.}''

\textbf{With feedback:}
``\textit{placed the metal lid onto the \underline{pot on the right side} of the stove, \underline{covering it}. The action has shifted from stirring or preparing to actively covering the pot, likely to let its contents simmer or cook.}''
\end{quote}

\noindent
The prior from Burst~1, where the model attended heavily to the pot and its contents, persists into Burst~2 and anchors the lid to the correct object. The correction comes from the model's own attention, not any outside signal, and the description now infers intent rather than only reporting position.

\begin{figure}[h]
    \centering
    \includegraphics[width=0.8\linewidth]{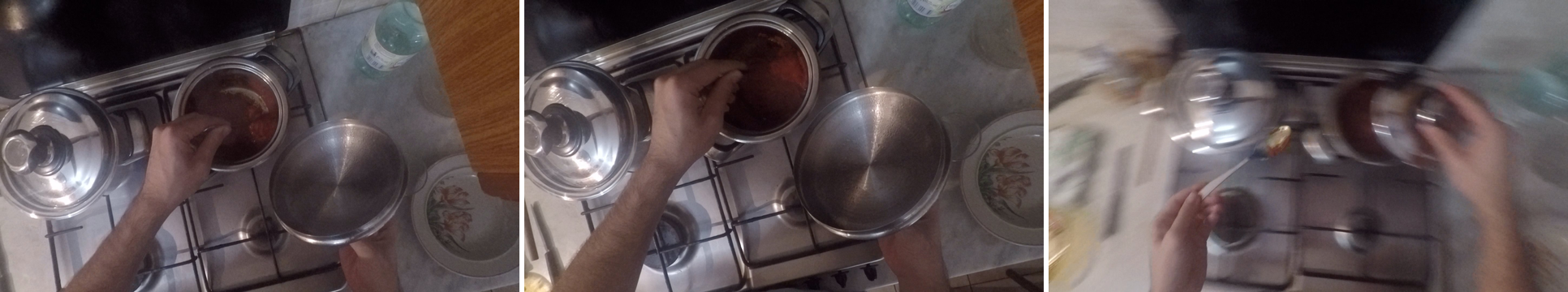}
    \caption{Frames from the sauce-stirring clip (EPIC-Kitchens, 25s).}
    \label{fig:sauce_frames}
\end{figure}

\paragraph{Feedback produces more focused descriptions.}
In a 25-second clip of stirring and transferring red sauce (Figure~\ref{fig:sauce_frames}), the third-burst descriptions are:

\begin{quote}
\small
\textbf{Without feedback} (5.5s inference):
``\textit{The person has now lifted the spoon \ldots\ poised to transfer the sauce. The spoon is suspended above the bowl, indicating the action of pouring is about to begin. ... The sauce in the pot appears to be thick and reddish-brown, and the bowl is empty and ready to receive the contents.}''

\textbf{With feedback} (3.2s inference):
``\textit{The person has now lifted the spoon \ldots\ poised to transfer the sauce.}''
\end{quote}

\noindent
By the third burst the prior has converged on the spoon--bowl--pot region, so the model stops re-describing layout it has already covered and the burst is 41\% faster. Uniform sampling here spends 15~bursts and 233s on 25 seconds of video ($\rho = 9.3\times$), 13 opening identically, against 3~bursts and 11s ($\rho = 0.44\times$) for FOCUS.

\section{Conclusion}
\label{sec:conclusion}
We presented FOCUS, a training-free framework that reduces VLM video inference cost
by gating frames in embedding space, steering each call with a spatial prior, and refining
that prior through closed-loop feedback. Under a matched backbone it holds the accuracy
of the strongest streaming baselines on EgoSchema at under half their VLM calls, and the
quality of the best captioning baseline at two-thirds of its compute ratio. The gain is not
better answers, but the same answers for substantially less computation. Appendix~\ref{app:limitations} gives
limitations and next steps.

\bibliography{colm2026_conference}
\bibliographystyle{colm2026_conference}


\appendix

\section{Notation}
\label{app:notation}

Table~\ref{tab:notation} collects every symbol used in Section~\ref{sec:method}.

\begin{table}[h]
\centering
\caption{Symbols used in Section~\ref{sec:method}.}
\label{tab:notation}
\smallskip
\footnotesize
\setlength{\tabcolsep}{3pt}
\begin{tabular}{@{}ll@{}}
\toprule
\textbf{Symbol} & \textbf{Meaning} \\
\midrule
$t$ & index of a frame in the arriving stream \\
$\mathrm{ref}$ & the reference frame for dissimilarity: the previously processed frame, updated every step \\
$g$ & vision encoder used for gating (Qwen2-VL-2B ViT) \\
$i$ & index of a spatial patch within a frame, $i \in \{1,\dots,P\}$ \\
$P, H_g, W_g$ & number of patches after merging, and the patch-grid height and width, $P = H_g W_g$ \\
$D$ & patch-embedding width \\
$\mathbf{E}^{(t)}, \mathbf{e}^{(t)}_i$ & patch-embedding grid of frame $t$ and the embedding of its patch $i$ \\
$d_i^{(t)}, D^{(t)}$ & per-patch cosine dissimilarity and its mean over patches \\
$\tau$ & gating threshold on $D^{(t)}$ \\
$B$ & burst size (frames captured per trigger) \\
$N_{\max}$ & safety timeout: forced capture after this many consecutively skipped frames \\
$n$ & burst index \\
$\mathbf{S}^{(t)}$ & per-patch dissimilarity reshaped to the $H_g \times W_g$ grid \\
$\mathbf{M}_{\mathrm{dissim}}$ & bottom-up dissimilarity map, max-pooled over the frames of a burst \\
$\mathbf{M}_{\mathrm{attn}}$ & top-down attention map fed back from the previous burst \\
$\gamma, \varepsilon$ & fusion exponent and floor in Eq.~\ref{eq:focus_prior} \\
$\mathbf{C}, C_j$ & fused spatial focus prior and its value at vision token $j$ \\
$S$ & decoder input sequence length \\
$j$ & index of a position in the decoder sequence \\
$\mathcal{V}$ & set of vision-token positions (input id equals the image-token id) \\
$\mathcal{T}$ & set of text positions \emph{after} the vision block, $\{j \notin \mathcal{V} : j > \max\mathcal{V}\}$ \\
$\ell, L$ & decoder layer index and number of layers ($L = 28$ for Qwen3-VL-8B) \\
$h$ & attention-head index \\
$\mathbf{V}^{(\ell)}_j$ & value-cache entry of position $j$ at layer $\ell$ \\
$s_j$ & per-step value-cache scale factor for vision token $j$ (Eq.~\ref{eq:value_scaling}) \\
$\alpha_f, \beta$ & focus (amplification) strength and suppression strength \\
$[r_{\mathrm{lo}}, r_{\mathrm{hi}}]$ & focus range: priors inside are amplified, priors outside are suppressed \\
$l_{\mathrm{start}}, l_{\mathrm{end}}$ & first and last calibrated decoder layer \\
$N$ & number of tokens generated in a burst response \\
$\mathbf{A}^{(\ell,h)}, \bar{\mathbf{A}}$ & raw attention tensor of layer $\ell$, head $h$, and its head/layer aggregate \\
$N_a$ & number of final decoder layers aggregated for attention extraction \\
$W$ & rolling window of previous responses carried forward as textual context \\
$\rho$ & VLM compute ratio (Eq.~\ref{eq:efficiency}) \\
$t_{\mathrm{vlm}}^{(n)}, t_{\mathrm{enc}}$ & VLM inference time for burst $n$; per-frame gating-encoder time \\
\bottomrule
\end{tabular}
\end{table}

\section{Full Algorithm}
\label{app:algorithm}

Algorithm~\ref{alg:focus} states the complete streaming procedure, covering frame
intake, the gating state machine, burst finalisation, spatial-prior fusion,
value-cache calibration, attention extraction, and prior persistence.

\begin{algorithm}[h]
\caption{FOCUS: streaming inference with closed-loop attention feedback}
\label{alg:focus}
\begin{algorithmic}[1]
\Require frame stream $\{x^{(t)}\}$; encoder $g$; VLM $f$; parameters $\tau, B, N_{\max}, \alpha_f, \beta, r_{\mathrm{lo}}, r_{\mathrm{hi}}, \gamma, \varepsilon, N_a, W$
\State $\text{state} \gets \textsc{capture}$;\ \ $\mathcal{Q} \gets \varnothing$;\ \ $\mathbf{M}_{\mathrm{attn}} \gets \varnothing$;\ \ $\mathcal{H} \gets \varnothing$;\ \ $k \gets 0$
\For{each arriving frame $x^{(t)}$}
  \State $\mathbf{E}^{(t)} \gets g(x^{(t)})$ \Comment{gating encoder runs on every frame}
  \State $d_i^{(t)} \gets 1 - \cos(\mathbf{e}^{(t)}_i, \mathbf{e}^{(\mathrm{ref})}_i)$;\ \ $D^{(t)} \gets \tfrac{1}{P}\sum_i d_i^{(t)}$;\ \ $\mathrm{ref} \gets t$
  \If{$\text{state} = \textsc{watch}$}
      \State $k \gets k + 1$
      \If{$D^{(t)} > \tau$ \textbf{or} $k \geq N_{\max}$} \State $\text{state} \gets \textsc{capture}$;\ $k \gets 0$ \Else{} \State \textbf{continue} \EndIf
  \EndIf
  \State $\mathcal{Q} \gets \mathcal{Q} \cup \{(x^{(t)}, \mathbf{d}^{(t)})\}$
  \If{$|\mathcal{Q}| < B$} \State \textbf{continue} \EndIf
  \Statex \hspace{\algorithmicindent}\textit{--- burst complete: run one VLM call ---}
  \State $\mathbf{M}_{\mathrm{dissim}} \gets \mathrm{norm}(\max_{t \in \mathcal{Q}} \mathbf{S}^{(t)})$ \Comment{Section~\ref{sec:value_cache}}
  \If{$\mathbf{M}_{\mathrm{attn}} \neq \varnothing$}
      \State $\mathbf{C} \gets \mathrm{norm}\!\big(\mathbf{M}_{\mathrm{dissim}} \odot (\varepsilon + \mathbf{M}_{\mathrm{attn}}^{\gamma})\big)$;\ \ install value-cache hooks with $s_j$ from Eq.~\ref{eq:value_scaling}
  \Else
      \State \textbf{no calibration} \Comment{first burst: orienting phase}
  \EndIf
  \State $y \gets f(\mathcal{Q},\ \text{prompt} \parallel \text{last } W \text{ responses of } \mathcal{H})$;\ \ remove hooks;\ \ $\mathcal{H} \gets \mathcal{H} \cup \{y\}$
  \State $\mathbf{M}_{\mathrm{attn}} \gets \textsc{ExtractAttention}(f, \mathcal{Q}, y)$ \Comment{Eqs.~\ref{eq:attention_aggregation}--\ref{eq:t2v_pool}; Appendix~\ref{app:attention}}
  \State $\mathcal{Q} \gets \varnothing$;\ \ $\text{state} \gets \textsc{watch}$;\ \ $k \gets 0$
\EndFor
\State \Return $\mathcal{H}$ \Comment{burst captions; for QA, one final call over $\mathcal{H}$ and the answer choices}
\end{algorithmic}
\end{algorithm}

\section{Attention Extraction in Detail}
\label{app:attention}

\textsc{ExtractAttention} runs once per burst, immediately after the burst
response has been generated, and returns the map $\mathbf{M}_{\mathrm{attn}}$
that becomes the top-down prior for the next burst.  The steps are:

\begin{enumerate}
\itemsep0pt
\item \textbf{One extra forward pass.}  Re-run the VLM over the concatenation of
  the burst prompt and the generated response with attention outputs enabled.
  This is a single teacher-forced pass with no autoregressive loop, so its cost
  is a small fraction of the generation it follows.
\item \textbf{Head and layer aggregation.}  Stack the last $N_a = 4$ layers,
  take the elementwise maximum over heads, then the mean over layers
  (Eq.~\ref{eq:attention_aggregation}).  Max-pooling over heads keeps the most
  spatially selective head instead of averaging it away with heads that attend
  almost entirely to text.
\item \textbf{Span selection.}  Partition positions into vision positions
  $\mathcal{V}$ (input id equal to the image-token id) and text positions, and let
  $\mathcal{T} = \{\, j \notin \mathcal{V} : j > \max \mathcal{V} \,\}$ be the text
  positions lying strictly after the vision block, that is the response tokens
  and any trailing prompt tokens.  Positions before the
  vision block are dropped because the causal mask gives them exactly zero
  attention to vision tokens, so retaining them would add rows of zeros and
  dilute the map.  If no post-vision text position exists, fall back to all text
  positions.
\item \textbf{Text-token reweighting.}  Write
  $\mathbf{A}^{\mathrm{t2v}} = \bar{\mathbf{A}}[\mathcal{T}, \mathcal{V}]$ and
  collapse it over text tokens by row-sum weighting rather than a plain mean,
  so a token that directs more total attention at the image contributes
  proportionally more:
  \begin{equation}
  \label{eq:t2v_pool}
  w_k = \frac{\sum_j \mathbf{A}^{\mathrm{t2v}}_{kj}}{\sum_{k'}\sum_j \mathbf{A}^{\mathrm{t2v}}_{k'j}},
  \qquad
  a_j = \sum_{k \in \mathcal{T}} w_k\,\mathbf{A}^{\mathrm{t2v}}_{kj}.
  \end{equation}
  Row sums are clamped below at $10^{-8}$ before normalisation.
\item \textbf{Grid folding.}  Split the $|\mathcal{V}|$ scores into their
  per-frame blocks using the processor's \texttt{image\_grid\_thw} metadata,
  reshape each block into that frame's own $(H_g^{(i)}, W_g^{(i)})$ grid, and,
  when dynamic resolution makes those grids differ, bilinearly resize each to the
  common $(H_g, W_g)$.  Average over the $B$ frames of the burst.
\item \textbf{Normalisation.}  Min--max rescale the result to $[0,1]$; if the map
  is constant, return zeros, which makes the next burst fall back to the
  dissimilarity signal alone.
\end{enumerate}

\section{Hyperparameters, Development Set, and Sensitivity}
\label{app:hparams}

\paragraph{Development set and tuning protocol.}
Every default in Table~\ref{tab:hparams} was fixed on a development set of
egocentric clips drawn from the same source datasets as the captioning benchmark
(Ego4D, EPIC-Kitchens, Ego-Exo4D and teleoperation recordings) but disjoint from
it: no video used for tuning appears in the 25-clip captioning benchmark or in
the 500-clip EgoSchema subset.  Selection was done once, by inspecting caption
quality and $\rho$ on the development clips, before any evaluation run, and no
value was revised after seeing evaluation results.  Two independent pieces of
evidence support that the resulting operating point is stable rather than fit to
a particular set.  First, the threshold sweep in
Table~\ref{tab:threshold_ego} is broadly flat in quality across a wide band of
$\tau$, so the operating point is not a sharp peak of the kind that leaks tuning
signal.  Second, the component ablation in Table~\ref{tab:components}, run on a
cohort sampled from EgoSchema and never used for tuning, selects the same
$(\alpha_f, \beta)$ setting that the development set did.

\paragraph{Hyperparameter inventory.}
The method carries a substantial number of hand-set constants, and we would
rather state that cost than bury it.  Table~\ref{tab:hparams} lists every one of
them, what it controls, and what we know about its sensitivity.  Two, $\tau$ and $B$, are swept
over a range; $\alpha_f$ and $\beta$ are toggled on and off in the component
ablation; the remainder were set once and never tuned per domain.

\begin{table}[h]
\centering
\caption{Complete hyperparameter inventory.  ``Swept'' indicates whether the
paper reports a sweep for that parameter.}
\label{tab:hparams}
\smallskip
\footnotesize
\setlength{\tabcolsep}{4pt}
\begin{tabular}{@{}llc>{\raggedright\arraybackslash}p{0.56\textwidth}@{}}
\toprule
\textbf{Symbol} & \textbf{Default} & \textbf{Swept} & \textbf{Role and sensitivity} \\
\midrule
$\tau$ & $0.4$ & yes & Gating threshold; trades compute against quality, flat over a wide band \\
$B$ & $32$ & yes & Burst size; quality within $0.07$ across $8$--$64$, $\rho$ minimised at $32$ \\
$N_{\max}$ & $300$ & no & Forced-capture timeout; liveness guard, inactive on all evaluated clips \\
$\alpha_f$ & $1.05$ & partly & Amplification strength; ablated on/off in Table~\ref{tab:components} \\
$\beta$ & $0.04$ & partly & Suppression strength; ablated on/off in Table~\ref{tab:components} \\
$[r_{\mathrm{lo}}, r_{\mathrm{hi}}]$ & $[0.75,1.0]$ & no & Focus range; what fraction of tokens is amplified vs.\ suppressed \\[-1pt]
floor & $0.5$ & no & Lower bound on the \emph{per-step} suppression factor; never binds at $\beta = 0.04$ \\
$\gamma$ & $0.5$ & no & Fusion exponent; how sharply attention modulates dissimilarity \\
$\varepsilon$ & $0.01$ & no & Fusion floor; keeps low-attention regions non-zero \\
$l_{\mathrm{start}}, l_{\mathrm{end}}$ & $0, L{-}1$ & no & Calibrated layer range; full stack by default \\
$N_a$ & $4$ & no & Final decoder layers aggregated for attention extraction \\
$W$ & $5$ & no & Rolling response window; identical for all methods and ablations \\
\bottomrule
\end{tabular}
\end{table}

\paragraph{Value-cache implementation notes.}
Three details of Eq.~\ref{eq:value_update} matter for reproducing it.  First,
the update is applied by a forward pre-hook that fires only when the incoming
hidden state has sequence length $1$, that is on decode steps and not during
prefill, so the values written by prefill are left alone and only their re-use
during generation is biased.  Second, the hook multiplies the persistent cache
in place, which is why the per-step factors compound to $s_j^N$ over an
$N$-token response; the magnitudes this reaches are given in
Section~\ref{sec:value_cache}.  Third, $s_j$ is computed once per burst rather
than once per token, so calibration costs one elementwise multiplication per
calibrated layer per decode step, which is negligible beside generation itself.

\paragraph{On reducing the hyperparameter count.}
Three of these are, in principle, removable rather than merely tunable.  $\tau$
is the most consequential and the most clearly replaceable: it is a fixed
threshold on a distribution ($D^{(t)}$ over a video) that the system already
observes online, so calibrating it to a running quantile of that distribution
would make gating adaptive at no extra encoder cost.  $[r_{\mathrm{lo}},
r_{\mathrm{hi}}]$ plays the same role in the spatial domain and could likewise be
set from a quantile of $\mathbf{C}$ rather than fixed.  $N_{\max}$ is a liveness
guard whose value never binds in our experiments.  The remaining constants are
either soft ($\gamma, \varepsilon$, the floor) or structural ($N_a$, $W$, the
layer range).  We flag adaptive thresholding as the highest-value follow-up in
Appendix~\ref{app:limitations}.

\section{Component and Threshold Ablations on EgoSchema}
\label{sec:ablations_main}

We ablate two axes: which components contribute what, and sensitivity to $\tau$.
Both use 50 videos randomly sampled from the 500-video subset, used only for
ablations, and both sweeps are single-run and should be read directionally.

\begin{table}[t]
\centering
\begin{minipage}[t]{0.545\textwidth}
\centering
\caption{Component ablation, 50 EgoSchema videos.  A dash means inactive.  Row (a)
is FOCUS (gating only) of Table~\ref{tab:egoschema}, row (c) is FOCUS (w/ Feedback)
as deployed, and (b) is an intermediate with amplification but no suppression.}
\label{tab:components}
\smallskip
\footnotesize
\setlength{\tabcolsep}{3.4pt}
\begin{tabular}{@{}lccccc@{}}
\toprule
\textbf{Configuration} & $\boldsymbol{\alpha_f}$ & $\boldsymbol{\beta}$ & \textbf{MCQ} & \textbf{Judge} & \textbf{Obj.} \\
\midrule
(a) Gating only        & --   & --   & 76.0\% & 3.81 & 3.76 \\
(b) + amplification    & 1.05 & --   & 76.0\% & \textbf{3.84} & \textbf{3.96} \\
(c) + suppression      & 1.05 & 0.04 & \textbf{78.0\%} & 3.81 & 3.92 \\
\bottomrule
\end{tabular}
\end{minipage}\hfill
\begin{minipage}[t]{0.43\textwidth}
\centering
\caption{Gating threshold sweep, same 50 videos.  VLM s/vid is mean VLM seconds
per clip; $\rho$ is against the 180\,s clip.  $\star$ marks the default.}
\label{tab:threshold_ego}
\smallskip
\footnotesize
\setlength{\tabcolsep}{3.4pt}
\begin{tabular}{@{}ccccc@{}}
\toprule
$\boldsymbol{\tau}$ & \textbf{MCQ} & \textbf{Judge} & \textbf{VLM s/vid} & $\boldsymbol{\rho}$ \\
\midrule
0.20 & \textbf{82.0\%} & \textbf{3.88} & 167.5 & 0.93 \\
0.30 & 74.0\% & 3.81 & 157.6 & 0.88 \\
0.40$^{\star}$ & 78.0\% & 3.81 & 131.7 & 0.73 \\
0.45 & 76.0\% & 3.81 & 96.7  & 0.54 \\
0.50 & 70.0\% & 3.81 & 54.7  & \underline{0.30} \\
\bottomrule
\end{tabular}
\end{minipage}
\end{table}

\paragraph{Gating versus calibration.}
Gating accounts for the entire reduction in calls; calibration supplies the quality
steering at the cost of one extra forward pass per burst.  Calibration is also not
separable from feedback when streaming: the cache is steered by $\mathbf{C}$, which the
closed-loop signal produces, so there is no ``calibration without feedback'' to ablate
(Appendix~\ref{app:offline} has the one offline case where it exists).  What varies
instead is calibration \emph{strength}, as Table~\ref{tab:components} does.  The clearest
effect is on object grounding, rising from $3.76$ to $3.96$ once amplification is
enabled (a\,$\rightarrow$\,b) and then holding, consistent with the prior concentrating
on the manipulated object; MCQ is flat from (a) to (b) and rises to $78.0\%$ once
suppression is added, while judge overall moves
only $3.81 \rightarrow 3.84 \rightarrow 3.81$.  At $n{=}50$ these deltas are small relative to what the cohort can resolve, so the table supports that calibration improves
object grounding without measurable degradation elsewhere, not that each
sub-component gives an independent gain.  The rolling text window is not ablated: it is
enabled identically for every baseline and variant.

\paragraph{Threshold sensitivity.}
$\tau$ trades compute against marginal quality rather than acting as a
quality-critical knob: judge quality is flat at $3.81$--$3.88$ across
Table~\ref{tab:threshold_ego} while $\rho$ falls threefold.  We choose $\tau = 0.4$
as the knee: it cuts VLM seconds by $21\%$ against the accuracy maximum at
$\tau = 0.2$ ($82.0\%$) while staying within $0.07$ judge and $4$ MCQ points of it,
and is explicitly \emph{not} that maximum.  Larger values buy compute at a visible
accuracy cost.  The per-domain breakdown below finds no domain with
a systematically different knee, though one whose dissimilarity distribution differs
materially from ours would need a different $\tau$.

\section{Per-Domain Threshold Sensitivity}
\label{app:domain}

Table~\ref{tab:domain} breaks the threshold sweep of
Table~\ref{tab:threshold_ego} down by activity domain.  Domain labels are derived
from the EgoSchema ground-truth caption summaries, since the subset carries no
official scenario tags.  Per-domain $n$ is small, so these are trends rather
than precise per-cell estimates: at $n \approx 10$ a single video is worth
$7$--$17$ percentage points.

\begin{table}[h]
\centering
\caption{Accuracy by domain across the gating threshold sweep (50 EgoSchema
videos).  ``Best'' is the highest accuracy that domain reaches at any $\tau$ in
the full sweep, and ``@0.40'' repeats the accuracy at our default.}
\label{tab:domain}
\smallskip
\small
\begin{tabular}{@{}lcccccc@{}}
\toprule
\textbf{Domain} & $\boldsymbol{n}$ & $\boldsymbol{\tau{=}0.30}$ & $\boldsymbol{\tau{=}0.40}$ & $\boldsymbol{\tau{=}0.50}$ & \textbf{Best} & \textbf{@0.40} \\
\midrule
Cooking \& food prep      & 14 & 79\% & 86\%  & 79\% & 86\%  & 86\% \\
Crafts, art \& textiles   & 11 & 64\% & 64\%  & 55\% & 73\%  & 64\% \\
Construction \& workshop  & 9  & 89\% & 89\%  & 89\% & 100\% & 89\% \\
Housework \& cleaning     & 6  & 83\% & 100\% & 83\% & 100\% & 100\% \\
Leisure, games \& social  & 10 & 60\% & 60\%  & 50\% & 70\%  & 60\% \\
\midrule
All (pooled)              & 50 & 74\% & 78\%  & 70\% & 82\%  & 78\% \\
\bottomrule
\end{tabular}
\end{table}

No domain exhibits a systematically different knee.  $\tau = 0.4$ lands at or
near each domain's best accuracy, and it is the best observed setting outright
for cooking and for housework.  Where it is not (crafts, leisure), the gap to
that domain's best is one or two videos wide at $n \approx 10$, which is within
sampling noise for these cell sizes.  We would still expect a domain whose
inter-frame dissimilarity distribution differs materially from egocentric
activity video, for instance static surveillance footage or fast-cut edited
video, to need a different $\tau$; that is the case adaptive thresholding is
meant to cover.

\section{Ablations on the Captioning Benchmark}
\label{sec:ablations_appendix}

The sweeps in this section use the 25-video captioning benchmark with GPT-5.2
judging the VLM responses, and complement the EgoSchema-based sweeps in
Appendix~\ref{sec:ablations_main}.  Because $n{=}25$, they should be read as
directional.

\paragraph{Capture ratio is rate-dependent, not threshold-dependent.}
At the same $\tau = 0.4$, gating retains $27\%$ of processed frames on this benchmark
but $74\%$ on EgoSchema (Section~\ref{sec:egoschema}), a gap worth stating plainly
because it changes what ``skipping redundant frames'' means on each.  The cause is
the input rate.  The captioning videos are ingested at ${\sim}23$~fps on average, so
consecutive processed frames are ${\sim}44$\,ms apart and most pairs are near
duplicates whose embedding dissimilarity falls below $\tau$.  EgoSchema is decoded at
6~fps, so consecutive processed frames are $167$\,ms apart, much more of the scene
has changed between them, and $D^{(t)}$ exceeds $\tau$ far more often; this is also
why the $N_{\max}$ liveness timeout never binds there.  The consequence for our
claims is limited but real: on EgoSchema the saving comes from issuing fewer,
longer-gap bursts (25.4 per clip against 55 for uniform sampling at the same
cadence) rather than from discarding frames, since a VLM invocation is billed per
burst and not per frame.  The frame-level reading of gating therefore holds at high
ingest rates and weakens as the input is pre-decimated, and a deployment sampling at
6~fps should expect capture ratios closer to the EgoSchema figure.

\begin{figure}[h]
  \centering
  \includegraphics[width=0.72\linewidth]{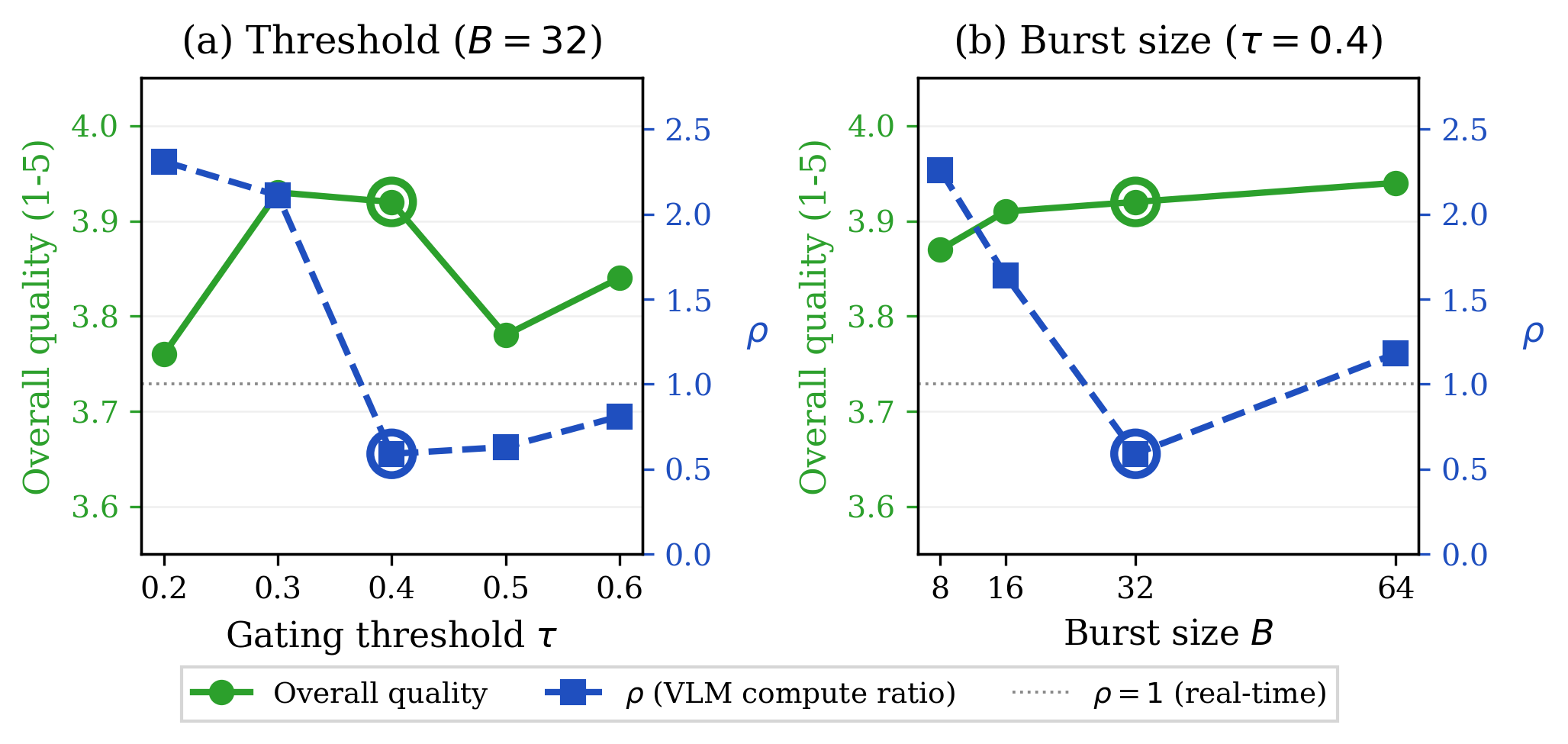}
  \caption{Quality and efficiency as a function of (a) gating threshold
  $\tau$ ($B{=}32$) and (b) burst size $B$ ($\tau{=}0.4$).  Circled markers
  denote the default configuration.  The dotted grey line marks $\rho{=}1$
  (real-time).}
  \label{fig:ablation_twopanel}
\end{figure}

\paragraph{Threshold.}
Quality peaks at $\tau = 0.3$ (3.93 overall) and remains within 0.01 at
$\tau = 0.4$ (3.92), before declining at higher thresholds.  The sharp drop in
$\rho$ between $\tau = 0.3$ and $\tau = 0.4$ (2.11 $\to$ 0.59) reflects the
distribution of inter-frame dissimilarity values in our videos: many
frame-to-frame transitions produce dissimilarity scores in the 0.3 to 0.4
range, so a small threshold increase crosses a critical point where
substantially fewer frames trigger bursts (capture drops from 65\% to 27\%).
Setting $\tau$ too low degrades quality because the VLM is overwhelmed with
redundant context; setting it too high ($0.5$+) skips important transitions.
We adopt $\tau = 0.4$ for its near-peak quality and the lowest $\rho$ in the
sweep, consistent with the EgoSchema-based sweep in
Table~\ref{tab:threshold_ego}.

\paragraph{Burst size.}
Smaller bursts ($B = 8, 16$) return to the \textsc{watch} state sooner, triggering more
VLM calls (17.6 and 12.3 on average).  Each call costs roughly the same
inference time (${\sim}7$s), so total VLM compute increases despite processing
fewer frames per call, yielding higher $\rho$.  The more frequent feedback
iterations explain why $B = 16$ achieves a higher Activity score (4.00 vs.\
3.96 at $B = 32$): the spatial prior converges faster with more updates.
Larger bursts ($B = 64$) give the VLM richer temporal context per call,
yielding the highest overall quality (3.94) and the best No Hallucination score
(2.72), at a moderate $\rho$ of 1.18.  The default $B = 32$ balances quality
(3.92) with the lowest $\rho$ (0.59) in the sweep.

\begin{table}[h]
\centering
\caption{Gating threshold ablation ($B = 32$, spatial + feedback enabled)
on the 25-video captioning benchmark.
Capture ratio is the fraction of input frames included in bursts.
$\dagger$\,denotes the default configuration used in Table~\ref{tab:main_results}.}
\label{tab:threshold_full}
\smallskip
\small
\begin{tabular}{@{}ccccccccc@{}}
\toprule
$\boldsymbol{\tau}$ & \textbf{Actor} & \textbf{Activity} & \textbf{Object} & \textbf{No Halluc.} & \textbf{Overall} & $\boldsymbol{\rho}$ & \textbf{Avg Calls} & \textbf{Capture} \\
\midrule
0.2 & 4.76 & 3.68 & 4.04 & 2.56 & 3.76 & 2.31 & 20.4 & 95\% \\
0.3 & 4.80 & 3.92 & \textbf{4.40} & 2.60 & \textbf{3.93} & 2.11 & 17.0 & 65\% \\
0.4\textsuperscript{$\dagger$} & \textbf{5.00} & \textbf{3.96} & 4.12 & \textbf{2.60} & 3.92 & \underline{0.59} & 8.9 & 27\% \\
0.5 & 4.76 & 3.80 & 4.04 & 2.52 & 3.78 & 0.63 & 5.2 & 12\% \\
0.6 & 4.84 & 3.88 & 4.08 & 2.56 & 3.84 & 0.81 & 5.2 & 12\% \\
\bottomrule
\end{tabular}
\end{table}

\begin{table}[h]
\centering
\caption{Burst size ablation ($\tau = 0.4$, spatial + feedback enabled)
on the 25-video captioning benchmark.
$\dagger$\,denotes the default configuration used in Table~\ref{tab:main_results}.}
\label{tab:burst_full}
\smallskip
\small
\begin{tabular}{@{}ccccccccc@{}}
\toprule
$\boldsymbol{B}$ & \textbf{Actor} & \textbf{Activity} & \textbf{Object} & \textbf{No Halluc.} & \textbf{Overall} & $\boldsymbol{\rho}$ & \textbf{Avg Calls} & \textbf{Capture} \\
\midrule
8  & 4.96 & 3.88 & 4.12 & 2.52 & 3.87 & 2.26 & 17.6 & 15\% \\
16 & 4.96 & \textbf{4.00} & 4.04 & 2.64 & 3.91 & 1.64 & 12.3 & 20\% \\
32\textsuperscript{$\dagger$} & \textbf{5.00} & 3.96 & \textbf{4.12} & 2.60 & 3.92 & \underline{0.59} & 8.9  & 27\% \\
64 & \textbf{5.00} & 3.92 & \textbf{4.12} & \textbf{2.72} & \textbf{3.94} & 1.18 & 7.2  & 36\% \\
\bottomrule
\end{tabular}
\end{table}

\section{LLM-as-Judge Protocol}
\label{app:judge}

GPT-5.2 scores each caption from 1 to 5 on four criteria: \emph{Actor Correct}
(is the agent correctly identified?), \emph{Activity Correct} (is the main
activity correct?), \emph{Object Correct} (are the key objects correct?) and
\emph{No Hallucination} (is the caption factually grounded?), with
\emph{Overall} their arithmetic mean.  To avoid penalizing stylistic variation
rather than substance, the judge is given explicit equivalences within actor
categories (``robotic arm'' $=$ ``robot hand'') and among activity synonyms
(``washing'' $=$ ``rinsing'').  The same prompt, criteria and equivalences are
used for every method, so judge bias applies uniformly across the comparison.  We
nonetheless treat judge-scored results as a per-criterion diagnostic only: our
headline quantitative claims come from EgoSchema, which is scored against
ground-truth answers.

\paragraph{Significance of the captioning margins.}
Because the quality differences in Table~\ref{tab:main_results} are small on 25
videos, we test them rather than reading them as ordered.  Each comparison is a
paired two-sided Wilcoxon signed-rank test over the 25 matched per-video
\emph{Overall} scores; $d_z$ is the mean paired difference divided by the standard
deviation of those differences.  Against BOLT, FOCUS (w/ Feedback) gives
$p = .895$ with $d_z = .03$; against DyCoke, $p = .158$ with $d_z = .32$; against
uniform sampling, $p = .256$ with $d_z = .24$; and against its own gating-only
variant, $p = .499$ with $d_z = .14$.  No planned comparison reaches $p < .05$, so
correcting for multiplicity does not change any conclusion, and we claim no quality
gain over the baselines.  Intervals below are percentile bootstrap over 100{,}000
per-video resamples at seed 2026.

\begin{center}
\footnotesize
\begin{tabular}{@{}lcc@{}}
\toprule
\textbf{Method} & \textbf{Overall} & \textbf{95\% CI} \\
\midrule
Uniform Sampling      & 3.81 & [3.66, 3.97] \\
DyCoke                & 3.81 & [3.60, 4.00] \\
BOLT                  & 3.91 & [3.74, 4.08] \\
FOCUS (gating only)   & 3.89 & [3.69, 4.07] \\
FOCUS (w/ Feedback)   & 3.92 & [3.73, 4.10] \\
\bottomrule
\end{tabular}
\end{center}

\noindent
The intervals overlap heavily, which is the honest summary of a 25-video study: the
efficiency separation is what this benchmark resolves, not the quality ordering.

\section{Offline (Non-Streaming) Sanity Check}
\label{app:offline}

Our claims concern the streaming setting, in which frames arrive in order and no
future frame or query is available.  BOLT and DyCoke were originally proposed for
the conventional offline setting, in which the entire clip is available upfront
and all selected frames are passed to the VLM in a single call.  To check that
the gating-plus-calibration signal is not an artefact of the streaming protocol,
we also ran both baselines in their native offline mode and built matching
offline variants of our method: \emph{FOCUS (offline)}, which gates the clip and
then passes the median frame of each retained burst to one QA call without
calibration, and \emph{FOCUS with calibration (offline)}, which adds a
dissimilarity-only spatial prior driving value-cache calibration.  Note that in a
single-pass offline setting there is no closed-loop feedback, because there is no
subsequent burst to feed attention back into; the prior is the bottom-up
dissimilarity map alone.  This is the one configuration in which calibration
exists without feedback, and it exists only offline.

In this regime our offline variants reach $74.4\%$ on the 500-video subset against
$74.0\%$ for BOLT (offline), so gating plus calibration transfers out of streaming.
That $74.4\%$ is a single pooled figure: we did not retain the two offline variants'
scores separately, and the DyCoke (offline) run is not tabulated here.  This appendix
therefore supports only the coarse conclusion that the signal survives the change of
protocol, and is not a like-for-like offline comparison.  We report it only as a
robustness check.  It is not a streaming result and is not part of the claim made in
Section~\ref{sec:egoschema}.

\section{EgoSchema Per-Answer Breakdown}
\label{app:egoschema_breakdown}

Table~\ref{tab:egoschema_breakdown} gives the per-answer breakdown for a single
representative run of FOCUS with feedback on the 500-video subset, the run that
scored $73.2\%$ ($366/500$).  We report a single run here rather than the 5-run
mean of Table~\ref{tab:egoschema} because the per-answer counts are only
meaningful within one run.

\begin{table}[h]
\centering
\caption{EgoSchema subset per-answer breakdown (500 videos, 5-way multiple
choice), single run of FOCUS with feedback in streaming mode, $\tau = 0.4$,
$B = 32$, 6~fps.}
\label{tab:egoschema_breakdown}
\smallskip
\small
\begin{tabular}{@{}lccccc|c@{}}
\toprule
 & \textbf{A} & \textbf{B} & \textbf{C} & \textbf{D} & \textbf{E} & \textbf{Overall} \\
\midrule
GT count   & 101 & 108 & 91 & 83 & 117 & 500 \\
Predicted  & 99  & 108 & 94 & 93 & 106 & 500 \\
Correct    & 72  & 79  & 70 & 67 & 78  & 366 \\
Accuracy   & 71.3\% & 73.1\% & 76.9\% & 80.7\% & 66.7\% & \textbf{73.2\%} \\
\bottomrule
\end{tabular}
\end{table}

The prediction distribution closely mirrors the ground-truth distribution,
indicating that the model does not exhibit strong answer-position bias.
Accuracy is relatively uniform across answer choices (66.7\%--80.7\%),
suggesting that performance is not dominated by surface-level shortcuts.

\section{Limitations and Future Work}
\label{app:limitations}

Gating is cheap only relative to the stream.  Where a 2B-scale encoder cannot
keep pace with arriving frames it, and not the VLM, becomes the bottleneck, and
the encoder term of Section~\ref{sec:frame_gating} governs rather than $\rho$.
The method carries twelve hand-set constants (Appendix~\ref{app:hparams}); we
sweep $\tau$ and $B$ and toggle $\alpha_f$ and $\beta$, leaving the rest
unexamined.  The captioning study is small ($n{=}25$) and LLM-judged, which is why
we lead with EgoSchema.  Most consequentially, gating is a decision not to look:
an event whose inter-frame dissimilarity falls below $\tau$ never triggers a
burst, and the timeout $N_{\max}$ bounds how long the system can go unobserving
without guaranteeing that any particular event is captured.  This is a property
of content-adaptive gating in general rather than an artefact of our thresholds,
and it bounds where a system of this kind belongs: we would not recommend it as
the sole perception path where a missed event carries safety consequences,
without an independent always-on channel for the events that matter.

The natural technical response to the last point is to stop fixing $\tau$ in
advance: it is a threshold on a distribution the system already observes online,
and Appendix~\ref{app:hparams} sets out how a running quantile would replace it
at no extra encoder cost.  Our other next steps are lighter gating encoders,
which lower that floor, and a multi-prior memory that would let the system carry
more than one region of interest across bursts.

\section{LLM Usage Disclosure}
\label{app:llm}

Large language models (GPT-5.2 and Claude) were used during the preparation of
this manuscript as assistive tools for: (1)~brainstorming paper structure and
experimental design, (2)~generating matplotlib plotting scripts,
(3)~rephrasing and copy-editing draft text, and (4)~converting plain-text
drafts to \LaTeX{} format.  All scientific claims, experimental results, and
technical contributions are the authors' own.  GPT-5.2 also serves as the
LLM-as-judge evaluator in our captioning benchmark (Section~\ref{sec:setup}),
which is an experimental component of the work, not a writing aid.

\end{document}